\documentclass{article}
\usepackage{amsmath,epsfig}
\usepackage{mathrsfs}
\usepackage{spconf}
\usepackage{algorithm}
\usepackage{algpseudocode}
\usepackage{mathrsfs}
\usepackage{bm}
\usepackage{stmaryrd}
\usepackage{graphicx}
\usepackage{subfigure}
\usepackage{amssymb}
\usepackage[table]{xcolor}
\usepackage{array,color}
\usepackage{verbatim}
\usepackage{booktabs}
\usepackage{ctable}
\usepackage{multirow}
\usepackage{multicol}
\usepackage{rotating}
\usepackage{threeparttable}
\usepackage[colorlinks=true,linkcolor=red,citecolor=blue]{hyperref}
\usepackage{cite}
\usepackage{enumerate}
\usepackage[titletoc]{appendix}
\usepackage[american]{babel}
\usepackage{microtype}

\title{Recent advances and opportunities in scene classification of aerial images with deep models}
\name{Fan Hu${}^{1,2}$, Gui-Song Xia${}^2$, Wen Yang${}^{1,2}$, Liangpei Zhang${}^2$}
\address{${}^1$ Electronic Information School, Wuhan University, Wuhan 430072, China\\
${}^2$ Key State Laboratory LIESMARS, Wuhan University, Wuhan 430072, China}
\begin{document}
%
\maketitle
\begin{abstract}
Scene classification is a fundamental task in interpretation of remote sensing images, and has become an active research topic in remote sensing community due to its important role in a wide range of applications. Over the past years, tremendous efforts have been made for developing powerful approaches for scene classification of remote sensing images, evolving from the traditional bag-of-visual-words model to the new generation deep convolutional neural networks (CNNs). The deep CNN based methods have exhibited remarkable breakthrough on performance, dramatically outperforming previous methods which strongly rely on hand-crafted features. However, performance with deep CNNs has gradually plateaued on existing public scene datasets, due to the notable drawbacks of these datasets, such as the small scale and low-diversity of training samples. Therefore, to promote the development of new methods and move the scene classification task a step further, we deeply discuss the existing problems in scene classification task, and accordingly present three open directions. We believe these potential directions will be instructive for the researchers in this field.
\end{abstract}
\begin{keywords}
Scene classification, datasets, deep models, scene caption, domain adaptation
\end{keywords}

\section{Introduction}
\label{sec:intro}

Nowadays, the new advanced remote sensing techniques for earth observation generate increasingly available high-resolution image data obtained from satellites, airplanes and unmanned aerial vehicles. The improvement of image resolution, resulting in more and more ground details clearly seen, calls for urgent development of automatic and accurate interpretation of numerous data sources. In this context, the classification of high-resolution remote sensing images turns from pixel- and object-level classification to scene-level semantic classification~\cite{yang2010bag,cheriyadat2014unsupervised}.

Scene classification of remote sensing images aims to categorize the given scene images into a set of semantically meaningful classes predefined according to human interpretation. Here, the \emph{scene} images refers to local patches extracted from large remote sensing images, containing specific semantic information, e.g., airport, park and residential area. Scene classification plays a significant role in a wide range of applications, e.g., urban planning and geographic mapping, and thus it has become a hot research topic in the remote sensing community.

Since the scene images usually cover multiple land-cover types or ground objects which vary in different scales, shapes, orientations and spatial distributions, scene images from different categories may share very similar content while images from the same category often exhibit high diversity in appearance. Such inter-class similarity and intra-class diversity make the scene classification a challenging task. Therefore, building robust and discriminative feature representations for describing the semantic content of scenes is the core component in the scene classification.

Over the past years, there has been growing interest in developing various methods for scene classification in remote sensing imagery~\cite{yang2010bag,CCMBOW,PSR,UFLSC,zhang2016scene,hu2015transferring,nogueira2017towards,Dodeepfeature,chaib2017deep,wang2017aggregating,li2017integrating}. From the pioneering work of introducing bag-of-visual-words (BoVW) model~\cite{yang2010bag,CCMBOW} to current state-of-the-art methods that employ the high-capacity deep convolutional neural networks (CNN)~\cite{AlexNet,VGGVD,he2016deep}, the performance of scene classification has gained dramatic breakthrough. However, despite the impressive results achieved by using modern CNNs, the accuracies on existing public datasets have almost reached saturation due to the limitations of these datasets. Recently, it seems that the studies on scene classification tend to blindly pursing accuracy improvement rather than providing enlightening ways to essentially address some problems still retained in scene classification task. Hence, in this paper, we would like to discuss some heuristic open directions for the sake of expanding the research scope of remote sensing image scene classification. We also wish these potential directions will arouse extensive attention from not only remote sensing community but also computer vision community. 
\section{A Brief Review on Scene Classification Methods}
The BoVW model is arguably the most popular approach in image classification applications~\cite{yang2010bag,CCMBOW}. It models an image as a histogram of visual words constructed by vector quantizing hand-crafted low-level features (e.g., the structural, textural or color features) with a clustering method. Due to the simplicity and good performance, the BOW model have also received many concerns in scene classification of remote sensing images. Motivated by the special spatial relationship among objects in scene images, several researchers have proposed improved variants of BOW by incorporating spatial information. However, the performance of BoVW and its variants is severely limited by the poor description ability of the hand-crafted features. To remedy this problem, several researchers resort to unsupervised feature learning (UFL) methods~\cite{cheriyadat2014unsupervised,hu2015transferring,zhang2015saliency}, which generate encoded features based on a codebook (also known as dictionary) learned from a large amount of unlabeled data. The UFL methods is capable of automatically learning discriminative features better adaptable to remote sensing scene images instead of relying on extraction of hand-crafted features.

More recently, the deep CNNs have enjoyed great success in various challenging visual recognition benchmarks, with substantially improving the state-of-the-art performance, owing to the large-scale image datasets and efficient use of high-performance GPUs~\cite{AlexNet,VGGVD,he2016deep,zhou2017places}. However, for a new visual recognition task with limited amount of training data, training a deep CNN that usually contains millions of parameters is infeasible. A lot of studies have discovered that intermediate features extracted from a deep CNN pre-trained on sufficiently large-scale dataset, such as ImageNet, can be employed as a generic image representation for many applications, e.g., scene classification, object detection and image retrieval. All these works focus on how to obtain strong image representations by transferring a pre-trained CNN to their tasks. For scene classification of remote sensing images, a few works~\cite{nogueira2017towards,castelluccio2015land} fine-tune the pretrained CNNs on the target scene datasets, and achieve promising performance; some works directly take the CNN activations from the fully-connected layers as image representations; the others~\cite{hu2015transferring,chaib2017deep,wang2017aggregating,li2017integrating} build discriminative feature representations by encoding CNN activations from convolutional layers in feature coding scenarios, where the convolutional features maps are viewed as a 2-D array of local features, obtaining the state-of-the-art performance on existing scene datasets. In general, the remarkable performance of those methods exploiting CNNs undoubtedly demonstrate the strong generalization capability of the pretrained CNN models.

To date, performance of CNN-based methods have gradually reach saturation, due to the serious limitations of existing datasets, such as the small amount of labeled training samples, the low-coverage and low-diversity. Hence, it is necessary to discuss more future directions in scene classification task so as to promote the development of understanding scenes in remote sensing imagery.

\section{Discussion on the Open Directions of Scene Classification}
\label{sec:OD}
In this section, we discuss several potential directions for scene classification in remote sensing imagery. They are inspired by the limitations of current scene classification methods and existing datasets. Here, we just present three examples.

\subsection{Building larger-scale datasets for scene classification}
Deep CNNs, as the most prevalent deep learning model, have demonstrated breakthrough accuracies and is now the dominant approach for almost all classification tasks. The availability of large-scale well-annotated datasets is one of the critical factors to the success of deep CNNs, because a considerable amount of training samples of wide coverage and great diversity are particularly beneficial to avoid overfitting and strengthen the generalization ability of CNN models. However, the existing datasets exhibit notable limitations, such as small number of image samples and scene categories, the less diversity within a categories, making it impracticable to fully train a deep CNN model from scratch. Because of these limitations, a majority of state-of-the-art methods resort to transferring deep CNNs that have been successfully pretrained on a large-scale natural image dataset (e.g., ImageNet~\cite{russakovsky2015imagenet}) for scene classification of remote sensing images, where the pre-trained CNNs are either considered as fixed feature extractors or fine-tuned (adjusting the parameters from a certain number of layers) on the target dataset. These kinds of transferring solutions perform fairly well on existing scene datasets with limited samples, but they are not optimal choice in comparison with training a deep CNN model from scratch, since fully trained deep networks is capable of learning more specific features perfectly adapting to the target datasets when the dataset is large enough~\cite{nogueira2017towards}. In other words, building large-scale datasets for scene classification is reasonably desirable, which can reach the full potentials of deep CNNs on one hand, and accelerate the widespread use of new-generation deep learning models on the other hand.

Very recently, it is noteworthy that tremendous efforts have been made to build large-scale benchmark datasets for scene classification., such as the AID~\cite{xia2017aid}, NWPU-RESISC45~\cite{cheng2017remote} and RSI-CB~\cite{li2017rsi}, the total number of samples being $10,000$, $31,500$, and $\sim36,000$ respectively. Compared with previous prevalent datasets, these new datasets occupy obvious superiority both in the total number and diversity of image samples. But so far, even the largest dataset is far too small to effectively train a deep CNN. A common deep CNN architecture with millions of parameters will dramatically overfit the tens of thousands of training samples. Therefore, a potential direction is to further enlarge the scale of current datasets to hundreds of thousands or even multiple millions.

Although it is convenient for us to get access to large amounts of remote sensing images with the rapid development of advanced remote sensing techniques, collecting image samples of good quality (here, ``good quality'' means the true category of each sample is purified and easily distinguished) is a very expensive and time-consuming task. It requires costly manual annotation from researchers with expert knowledge in the field of remote sensing image interpretation. Without any filed trips or survey, expertise and research experience are essential to ensure the accurate labeling.

Considering that it is challenging to well annotate a high-coverage and high-diversity dataset (time-consuming and call for expert domain knowledge), labeling scene images using geographic crowdsource data, e.g., the Open Street Map, is a feasible way, since the crowdsource data labeled by the public are massive in quantity, highly real-time and acquired in low cost. We expect that more researchers will focus on establishing larger-scale high-quality datasets for scene classification on the basis of the crowdsource data. The future large-scale dataset will in turn promote the community to develop new deep models specially appropriate for remote sensing scenes.

\subsection{Better describing the content of scenes}
The well-defined scene classification task is to label scene images with specific semantic categories. In other words, the \emph{terms} of semantic categories (e.g., commercial area and residential area) summarily describe the semantic content of scenes in high-level abstraction. Such kind of static classification task cannot fully describe the attributes of ground objects and their relationship in scenes. Thus, in order to show more detailed content in scene images, it is potential to move scene classification further to scene caption, which aims to describe the content of scenes in a finer semantic level using meaningful sentences. Thereby, a scene image is depicted by dynamic sentences summarizing the main content of scenes instead of a static single term of category.

The methods developed for natural image caption may not adapt well to scene caption in remote sensing images, due to ambiguous semantic information (objects show variations in scale, orientation and geometric structure) and complicated spatial distribution of objects. To follow the direction of scene caption, a well-annotated scene caption dataset is also necessary. Researchers have presented a few exemplary works on remote sensing image caption~\cite{lu2017exploring,shi2017can}, and have constructed a large-scale dataset under specific annotated instructions in consideration of characteristics of remote sensing images, e.g., not using words that represent the concept of ``\emph{direction}'' and ``\emph{vague}''. We believe that the scene caption will be a new chance to generate better description of scenes in remote sensing images and will receive more concerns from remote sensing community.

Besides, the online geo-tagged pictures collected from social websites and location-based geographic resources (e.g., the global positioning system) have great potentials for recognizing remote sensing scenes, since they are able to supply various information, e.g., finer details of all kinds of ground objects with extremely high resolution, vertical views rather than overhead views, where the ground objects are located and what happens at that time~\cite{cheng2017remote}. This kind of crowdsourcing information can be employed to train new models together with remote sensing dataset, and thus will be valuable for better understanding the content of remote sensing scenes. 

\subsection{Dealing with images from different domains}
As described above, the current studies have proven that the CNNs are valuable tools for constructing ``black box'' architectures to accurately interpret remote sensing image scenes. In particular, the successful pretrained CNNs generalise well to remote sensing scene datasets, and have been reported with impressive classification performance when used to build discriminative holistic representations from intermediate layers or to perform fine-tuning of model parameters on target datasets. However, these strategies of exploiting pretrained CNNs may be not effective enough for images acquired under different conditions (e.g., the labeled training images and target images are obtained from different sensors varying greatly in spatial and spectral resolutions). In fact, the remote sensing images are inevitably affected by various natural and human factors, such as the sensors, perspective of camera, season and weather conditions, geographic locations, and so on. Thus, the simple transferring strategies for pretrained CNNs are likely to achieve unsatisfactory results when the source dataset is far from similar to the target dataset (also called \textit{data shift}). This problem can be well addressed by more complex approaches concerning domain adaptation~\cite{othman2017domain}.

To facilitate the generalization power of these CNN models when dealing with data shift , it is a potential opportunity of designing improved approaches based on domain adaption, wherein the feature representations from target and source domain are properly mapped in a uniform space while preserving original structures of data. It is feasible to develop additional adaptation layers (or even well-designed network modules) based on the output of a pretrained CNN, and to optimize the loss function with imposing auxiliary regularization terms to reduce the mismatch between target and source data distributions~\cite{othman2017domain}.

\vspace{-2mm}
\section{CONCLUSION}
\label{sec:conclusion}
The development of scene classification methods of remote sensing images has hit a bottleneck in the context of revolutionary change caused by deep models. In this paper, we delve into the existing issues in scene classification, and put forward several open directions to broaden the scope of scene classification task. We believe these potential opportunities will receive increasing interests from both the remote sensing and computer vision community, and thereby will encourage the communities to develop new models.

{
\small
\bibliographystyle{IEEEbib}
\bibliography{hfbib}
}

\end{document}